\documentclass[conference]{IEEEtran}
\IEEEoverridecommandlockouts
\usepackage{cite}
\usepackage{amsmath,amssymb,amsfonts}
\usepackage[ruled,vlined]{algorithm2e}
\usepackage{CJKutf8}
\usepackage{multirow}
\usepackage{url}
\usepackage{graphicx}
\usepackage{textcomp}
\usepackage{xcolor}

\usepackage{booktabs}
\def\BibTeX{{\rm B\kern-.05em{\sc i\kern-.025em b}\kern-.08em
    T\kern-.1667em\lower.7ex\hbox{E}\kern-.125emX}}
\usepackage{diagbox}
\begin{document}

\title{\huge Blur the Linguistic Boundary: Interpreting Chinese Buddhist Sutra in English via Neural Machine Translation
}
\author{\IEEEauthorblockN{Denghao Li$^{\sharp}$$^{\dagger}$\thanks{$^{\dagger}$Equal contribution}, Yuqiao Zeng$^{\sharp}$$^{\diamond}$$^{\dagger}$, Jianzong Wang$^{\sharp}$$^{\star}$\thanks{$^{\star}$Corresponding author: Jianzong Wang, jzwang@188.com.}, Lingwei Kong$^{\sharp}$, Zhangcheng Huang$^{\sharp}$,\\ Ning Cheng$^{\sharp}$, Xiaoyang Qu$^{\sharp}$, Jing Xiao$^{\sharp}$}
\IEEEauthorblockA{\textit{$^{\sharp}$Ping An Technology (Shenzhen) Co., Ltd., China} \\ \textit{$^{\diamond}$University of Electronic Science and Technology of China, China}}
}

\maketitle

\begin{abstract}

Buddhism is an influential religion with a long-standing history and profound philosophy. 
Nowadays, more and more people worldwide aspire to learn the essence of Buddhism, attaching importance to Buddhism dissemination. 
However, Buddhist scriptures written in classical Chinese are obscure to most people and machine translation applications. 
For instance, general Chinese-English neural machine translation (NMT) fails in this domain. 
In this paper, we proposed a novel approach to building a practical NMT model for Buddhist scriptures. 
The performance of our translation pipeline acquired highly promising results in ablation experiments under three criteria.
\end{abstract}


\section{Introduction}
Buddhism is one of the three major religions worldwide founded by Siddhartha Gautama in ancient India between the 6th and 5th centuries BC. 
It has spread widely and significantly impacted the sociopolitical and cultural life of numerous countries. 
To propagate Buddhism, it is necessary to translate Buddhist scriptures into English to help more people understand the essence of Buddhism. 
Nowadays, the application of artificial intelligence technologies like neural machine translation (NMT) has been accelerating translation\cite{dabre2020survey,2020An}. 
However, the amount of Chinese-English parallel corpus in the Buddhism domain is limited, making Buddhist scriptures translation low-resource. 
In recent years, numerous researches on low-resource NMT have been proposed, which can be classified into three categories based on different source data types: monolingual data, multi-modal data, and data from auxiliary languages\cite{wang2021survey}.

Common methods exploiting monolingual data are back translation and unsupervised NMT.
Back translation is to generate synthetic bilingual corpus by substantial monolingual data \cite{epaliyana2021improving,zhou2021improving}. 
However, the quality of parallel data generated by back translation is usually not ideal. 
Unsupervised NMT utilizes monolingual data only to acquire primitive alignment and improves translation by iterative back translation \cite{ren2020retrieve,duan2020bilingual}. 
Nonetheless, existing unsupervised learning approaches can hardly reach a readable performance on long sequence translation. 

Parallel data in other modalities, such as image, video and speech, can also be applicable for NMT \cite{huang2021image,lekshmy2022english,zhao2022nnspeech}. 
Nevertheless, these approaches are not generalized enough and typically focus on specific circumstances, which cannot benefit most text-to-text tasks.

NMT can also get benefits from auxiliary languages which are rich-resource in the same language family of the language which needs to be translated. 
Many methods apply auxiliary languages to improve NMT performance, such as multilingual training and transfer training. 
In multilingual training, auxiliary language is trained with low-resource language\cite{zhang2021bidirectional,jwalapuram2020pronoun,weller2020modeling}, while in transfer learning trains model by auxiliary language and fine-tunes the existing model with low-resource languages\cite{artetxe2020translation,ji2020cross,2021Modeling}. 
Tuning from pre-trained models has been one of the mainstream approaches in multiple natural language processing tasks including translation. 
However, simply applying tuning to a pre-trained NMT model failed to fix the input distribution variance between classical and vernacular Chinese, and is not an efficient way to learn rare words in the specific domain.

This paper proposed tuning a pre-trained encoder with a designed loss item to adjust to Buddhist scriptures, leveraging the knowledge learned from vernacular Chinese. 
Concerning the model, memory information retrieval in the decoder served domain-specific Chinese-English translation.
The contributions of our work are listed as follows:

\begin{itemize}
    \item Corpus Collection: A Chinese-English parallel corpus on Buddhism was established, containing 90 volumes of scriptures.
    \item Low-Resource Solution: A vernacular Chinese was introduced as pre-training model. The inconsistent meanings between classical and vernacular Chinese expression were unified via side network tuning.
    \item Optimized Professionalism: A sequence-level proper nouns recall mechanism leveraging Aho-Corasick algorithm was designed to improve the professionalism of the system.
\end{itemize}

\section{Related Works}

Challenges of Buddhist scriptures translation mainly result from it being low-resource and domain-specific.
On the one hand, the parallel Chinese-English corpus of Buddhist scriptures is limited, namely low-resource. 
It is feasible to utilize the similarity between classical Chinese, recording Buddhist scriptures, and vernacular Chinese. 
However, there exists inconsistency between the meanings of classical Chinese and vernacular Chinese, even though both of them consist of Chinese characters. 

Ladder side-tuning (LST) \cite{sung2022lst} is an efficient transformer tuning architecture that flexibly combines pre-trained knowledge with tuning tasks\cite{zhang2020side,zingaro2021multimodal}. 
The ladder side network constructs a side connection block for each encoder block and propagates the gradients directly through them. 
Therefore, the tuning will not influence the value of weights in the backbone pre-trained model. 
In this way, each side connection block can be regarded as a pruned copy of the corresponding block in the backbone network, with fully connected projection layers at the beginning and end, making the vector dimension represented by tokens unchanged. 
The output of the backbone block is combined with that of the corresponding side connection block via a gated unit. 
The output weights of these two blocks are learned during training.

On the other hand, the Buddhist scriptures translation is domain-specific.
Substantial proper nouns about Buddhism are rare in the general corpus, making scriptures less understandable to human and machine translators. 
The following works contribute to the domain-specific translation.
Retrieval-based NMT \cite{khandelwal2020nearest} is an enhancement of standard transformer-based models. 
Token level k-nearest-neighbor retrieval \cite{wang2022efficient} has successfully introduced explicit decoding targets into NMT systems. 
In detail, a key-value datastore is created at the model training phase for target token retrieval. 
Let $S$ and $T$ be the source and target corpus, respectively. $s$ and $t$ denote a sample pair of parallel sequences. 
The keys $h(s, t_{<i})$ are representations of training target tokens $t_i$ decoded by a pre-trained auto-regressive NMT model h, and the values are token $t_i$. 
During the prediction phase, the output of the decoder will be searched in the datastore to obtain its k-nearest-neighbors and their corresponding tokens. 
The output token is generated from token sets mentioned above by specific strategies. 
For example, in \cite{zheng2021adaptive}, all possible tokens are weighted by applying a Gaussian kernel function.

\begin{figure*}[htbp]
\includegraphics[width=0.8\textwidth]{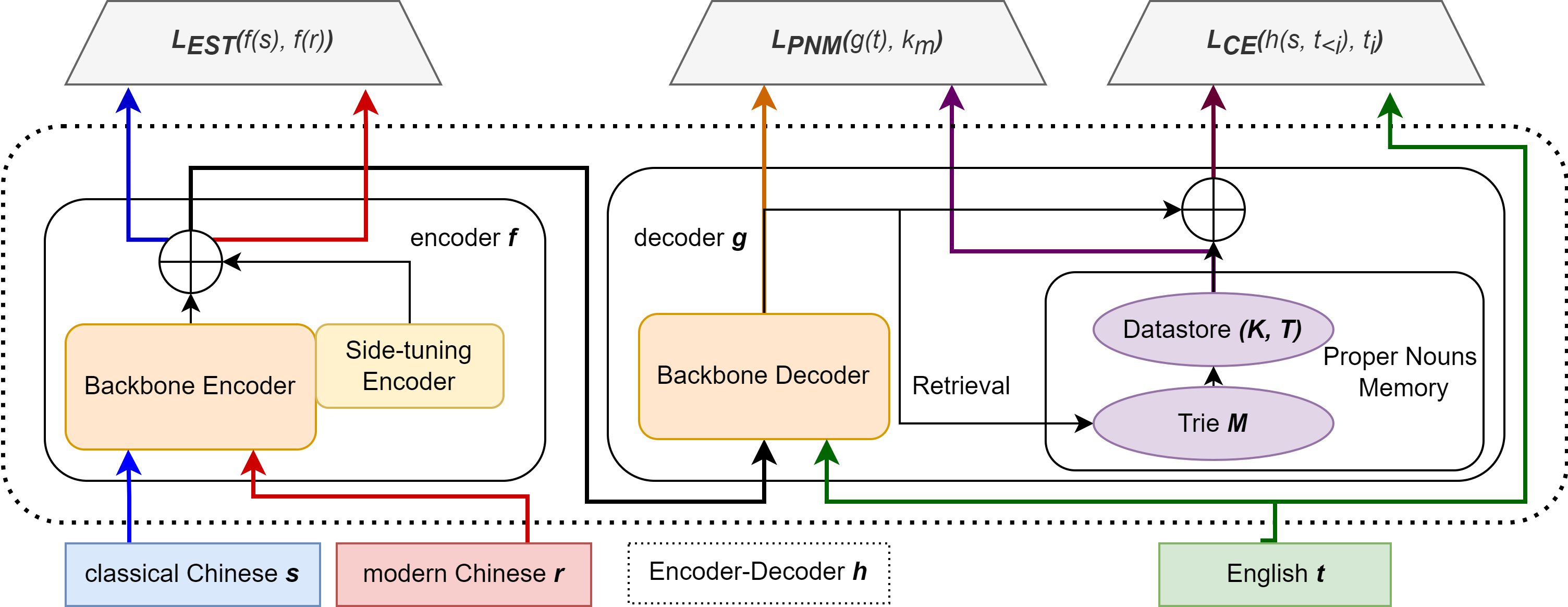}
\centering
\caption{Illustrations of our proposed approach: A side-tuning encoder (orange block) is trained to bridge the gap between classical and vernacular Chinese, and a memory (purple blocks) including a trie $M$ and a datastore $(K, T)$ is built to learn proper nouns. Two loss items are designed, along with the fundamental cross entropy (CE) loss (gray blocks), to optimize the system.}
\label{fig:method}
\end{figure*}

\section{Methodology}

Our proposed approach for scripture translation is illustrated in Figure~\ref{fig:method}. 
On the encoder side, we applied side-tuning to help the model understand classical Chinese based on its knowledge of vernacular Chinese. 
On the decoder side, we constructed the memory of the training corpus by forwarding the training data through a pre-trained model and mapping decoder output vectors to target tokens. 
We then utilized it to improve the performance of proper nouns. 

\subsection{Encoder Side-Tuning}
As scripture text shares the same token embedding as general corpus, it is possible to tune the encoder of the pre-trained translation model for scripture input. 
However, tuning techniques focusing on downstream layers after encoding hardly worked in the task. 
The tuning objective is encoding itself rather than any downstream tasks.
Thus the tuning process should be applied to the whole encoding process.

As an encoder-only task, the gradients should optimize a loss item generated based on the encoding quality. 
This loss is computed by evaluating the differences between encoding source sentences and reference sentences: 
\begin{equation}
    L_{EST} = \sum_{(s, t, r) \in (S, T, R)} L(f(s), f(r))
\end{equation}

where the loss function $L$ is the sum of cosine distance between
each pair of representations encoded by encoder $f$. 
And the reference sentences $R$ come from two resources, our collected vernacular Chinese version of the scriptures and back-translated results of target scriptures $T$.

\subsection{Proper Noun Memory}
As a domain-specific task, the translation of the proper nouns is essential. Namely, the model needs to train with a substantial corpus. 
However, the size of the parallel corpus in the Buddhist scripture domain is limited. 
Hence statistical instruction brought by gradient descent cannot lead the decoder to translate proper nouns correctly. 
Moreover, proper nouns usually consist of a variable-length sequence of tokens rather than a single token, making vanilla k-nearest-neighbor token retrieval fail to find the targets precisely.
Consequently, we designed a novel method to retrieve the proper nouns.

Firstly, we generalized the retrieval process to sequence-level with the Aho-Corasick algorithm. 
In detail, the proper noun memory construction is illustrated in Algorithm \ref{alg:0}.
Proper nouns are recognized from parallel corpus during the word-level tokenization procedure of the pre-processing with a part-of-speech tagging tool $POS$. 
Moreover, corresponding token sequences are recorded in a trie tree while applying byte pair encoding $BPE$.
Then in model forwarding, target tokens of the k-nearest-neighbor of decoder outputs will be searched by the pre-built Aho-Corasick automation.

\begin{algorithm}[ht!]
    \caption{Proper Noun Memory Building}
    \label{alg:0}
    \KwIn{
    parallel corpus $(S_{raw}, T_{raw})$;\\
    pre-trained encoder-decoder model $h$;}
    \KwOut{
    trie $M$;corpus $(S, T)$;datastore $(K, T)$;\\
    }
    
    {Initialize byte-pair-level trie $M$;\\}
    \For {$(s_{raw}, t_{raw}) \in (S_{raw}, T_{raw})$}
    {
    $(s_{tok}, t_{tok}, s^{PN}_{tok}, t^{PN}_{tok}) = POS(s_{raw}, t_{raw})$;\\
    $(s, t, s^{PN}, t^{PN}) = BPE(s_{tok}, t_{tok}, s^{PN}_{tok}, t^{PN}_{tok})$;\\
    \lIf{$t^{PN}_{i,...,j}$}{Add $t_{i,...,j}$ to trie $M$}
    Add $(s, t)$ to corpus $(S, T)$;\\
    \For {$(k_{i}, t_{i}) \in (h(s, t), t)$} 
    {Add $(k_{i}, t_{i})$ to datastore $(K, T)$;}
    }
\end{algorithm}

Let $\mathbb{I}_{PN}$ be the indicator function that masks the tokens of proper nouns.
The memory retrieval process is also driven by the loss item measuring the distances between the decoder $g$ output and the anchor $k_m$, which is the center of gravity of the stored token cluster:
\begin{equation}
    L_{PNM} = \sum_{(s, t) \in (S, T)} \sum_{(k_i, t_i) \in (K, T)}^{\mathbb{I}_{PN}(t_i)} L(g(t_i), k_m)
\end{equation}

\section{Experiments}

\subsection{Model Setup}

The pre-trained model for our Chinese-English translation task is the standard Transformer architecture implemented with fairseq \cite{ng2019facebook}. 

In detail, both encoder and decoder are set to have 6 layers. 
The number of attention heads is 8. 
The hidden embedding size is configured to be 512 and the feed-forward layer embedding size is 2048. 
Inverse square root learning rate decay policy is utilized with 4000 warm-up steps.
WMT 2019 Chinese-English dataset containing 25,986,436 samples is harnessed for training the pre-trained and baseline model.

We applied a four times down scaled transformer encoder for side tuning.
Each layer of the side tuning encoder takes the summation of last layer's output and backbone network output as input, where the backbone network output is the output of corresponding backbone network layer through a linear down scale layer.
During proper nouns memory building, tokenization and part-of-speech tagging were conducted with NLTK standard for English and CTB standard for Chinese respectively.
Then both corpus were byte pair encoded with subword-nmt toolkit.

\subsection{Buddhist Datasets}
\subsubsection{Data Sources}
The corpus we use is mainly the Tripitaka, of which there are 1,659 original texts of the Tripitaka, totaling 7,143 volumes.
There are 284 English scriptures and 133 Chinese scriptures. 
The specific information is shown in Table \ref{tab:1}. 
The total monolingual data collection contains approximately ten million words.

\begin{table}[!htbp]
    \centering
    \caption{Monolingual Data Sources Statistics}
    \begin{tabular}{c|c|c}
        \toprule
        Data source & Language 	& Scriptures number \\
        \midrule
        BDKAmerica 	& English	& 64       \\
        City of Ten Thousand Buddhas	& English	&  16	    \\ 
        84000 & English	& 204      \\ 
        Chinese Buddhist Website 	& Chinese	& 24  \\ 
        Xianmi Library 	& Chinese	& 109  \\ 
        \bottomrule
    \end{tabular}\label{tab:1}
\end{table}

\subsubsection{Corpus Preprocessing}

We firstly formulated the rules in data processing as follows:

\newtheorem{definition}{Definition}
\begin{definition}
When filtering the corpus content, we eliminated non-Buddhist scripture content like annotations and extracted the text content of Buddhist scriptures only.
\end{definition}

\begin{definition}
Sentence segmentation is according to Chinese and English punctuation marks (including comma, period, semicolon, exclamation marks, and question marks). The length of the single-segmented sentence should be at least five words.
\end{definition}

The corpus processing part has three levels of Chinese-English scriptures alignment: document-level, sentence-level, and word-level. 
After monolingual data collection, the number of matched scriptures is ten. They are \emph{The Buddha Speaks of Amitabha Sutra, Mahavairocana Tantra, Buddhavatamsaka Sutra, Sutra of the Terra Treasure, Diamond Sutra, The White Lotus of the Good Dharma, the three pure land of sutras, The Sutra in Forty-Two Sections said by Buddha, Heart Sutra, Pharmacist}, and, in total, 90 volumes.



The corpus is firstly filtered according to Definition 1. 
It is then segmented into single sentences, acquiring 21,961 single sentences in English and 50,525 single sentences in Chinese.
After sentence segmentation, the Chinese and English corpus are matched accordingly. We applied Language-agnostic BERT sentence embedding (LaBSE) \cite{feng2022language} to vectorize the sentences and match the sentences by maximum cosine similarity between the sentences in a shifting window along the whole scripture.



\subsubsection{Proprietary Glossary}
Buddhist scriptures differ from conventional articles for containing many proper nouns in Buddhism and unique nouns in classical Chinese, such as official positions, era names, etc. 
To deal with this problem, we firstly marked the proper nouns with part-of-speech tagging tools. 
We then checked the results by querying the electronic Buddhist dictionary to ensure the correct and professional translation. 
Buddhist vocabulary was constructed following the ``five principles of non-translatio'' in Buddhism. 
Finally, a set of proprietary vocabulary was built, including classical Chinese, vernacular Chinese, and English. 
The vocabulary was divided into five categories: proper nouns in Buddhist scriptures, place names, official positions, personal names, and others (time, year, etc.). 
The specific data are shown in Table \ref{tab:pn}.

\begin{table}[!htbp]
    \centering
    \caption{Proprietary Glossary Information Statistics}
    \begin{tabular}{c|c}
        \toprule
        Classification & Number	 \\
        \midrule
        Buddhist proprietary glossary & 362     \\
        Places	& 45	 \\ 
        Official positions & 11	 \\ 
        Persons 	& 97  \\ 
        Others 	& 10  \\ 
        \bottomrule
    \end{tabular}\label{tab:pn}
\end{table}

\subsection{Evaluation Metrics}
The most widely used evaluation metric in NMT is bilingual evaluation understudy (BLEU). 
In this work, we applied the BLEU calculator provided by the NLTK toolkit. 
Moreover, as a domain-specific translation task, it is essential to introduce other criteria, serving as a supplement. 
On the one hand, the accuracy of the proper nouns should be assessed meticulously. 
On the other hand, judgment from human beings, exceptionally professional Buddhist experts, is also necessary.

\subsubsection{Proper Noun Accuracy}
We rated the translation accuracy of the proper nouns via our preprocessing methods (PNA). 
The words are coped at the subword level by the BPE processing. 
The results are evaluated at a higher level. 
Specifically, the translation is reckoned as correct only if all of the proper nouns are correct, i.e., the whole token sequence is accurate at every position.

\subsubsection{Human Evaluation}
We also invited ten professional Buddhist practitioners to score the translations from 1 to 5, where 1 is the worst and 5 is the best. 
The scoring came from a weighted average of three aspects. The first is precision. 
The translation should reflect the meaning of the source texts precisely.
The second is fluency. 
The translation should be understandable without grammar errors or ambiguities, conforming to modern language habits. 
The third is elegance. 
Many sentences in original scriptures have particular rhythms like poems, and the translation is also expected to be of high literary value.

\section{Results}

\begin{table}[!htbp]
\centering
\caption{Evaluation Results on Test Set}
\begin{tabular}{c|c|c|c}
\toprule
\diagbox {Systems} {Metrics} & BLEU & PNA & Human \\
\midrule
Baseline & 13.65 & 53.70 & 4.16 \\
Baseline+EST & 16.48 & 54.25 & 4.27 \\
Baseline+PNM & 16.05 & 85.72 & 4.21 \\
Baseline+EST+PNM & 18.48 & 88.22 & 4.35\\
\bottomrule
\end{tabular}\label{tab:res}
\end{table}

The evaluation results are shown in table \ref{tab:res}.
We designed an ablation experiment to examine the performance of our model. As our model mainly consists of encoder side-tuning (EST) and proper noun memory (PNM), the experimental group setup is shown in Table \ref{tab:res}. Group 1 applied a transformer, serving as a baseline for this experiment. Group 2 added encoder side-tuning (EST) on the baseline, and Group 3 supplemented proper noun memory (PNM). Group 4, representing our model, added EST and PNM based on the transformer. 

Table \ref{tab:res} shows that the following statements are true under all three metrics. Group 1 (baseline) displays the lowest performance. Namely, both Group 2 (baseline+EST) and Group 3 (baseline+PNM) show better performance than Group 1, proving that these two blocks can optimize model performance, respectively. In particular, Group 4 (baseline+EST+PNM) performs the best among the four groups, verifying that our model improves translation quality to a great extent.

\subsection{Case Study}


\begin{table*}[!htbp]
\centering
\caption{Case Results on Buddhist Dictum and Classical Buddhist Scriptures}
\label{tab:case}
\begin{tabular}{l|l|l}
                 Buddhist scriptures & Methods & Results           \\ \hline
\multirow{4}{*}{\begin{CJK*}{UTF8}{gbsn}色即是空，空即是色。\end{CJK*} }
                  & Baseline & Color is empty, emptiness is color. \\
                  & EST & Form is void, and emptiness is form. \\
                  & PNM & Color is emptiness, emptiness is color. \\
                  & EST+PNM & Form is emptiness, and emptiness is form. \\ \hline
\multirow{4}{*}{\begin{CJK*}{UTF8}{gbsn}如来所得法，此法无实无虚。\end{CJK*}}
                  & Baseline & Tathagata's income method, this method is not real and not virtual. \\
                  & EST & The dharma obtained by Tathagata, this dharma has no reality or emptiness.  \\
                  & PNM & Tathagata's income method, this method is not real and not virtual. \\
                  & EST+PNM & The dharma obtained by Tathagata, this dharma has no reality or emptiness. \\ \hline
\multirow{4}{*}{\begin{CJK*}{UTF8}{gbsn}\shortstack{长老须菩提，及诸比丘，\\{比丘尼，优婆塞，优婆夷}}\end{CJK*}}
                  & Baseline & The elder must Bodhi, and all bhikkhus, bhikkhunis, youposai, youpoyi. \\
                  & EST & The elder Subodhi, and all the bhikkhus, bhikkhunis, youposai, youpoyi. \\
                  & PNM &  The elder must Bodhi, and all bhikkhus, bhikkhunis, upasakas, upasikas. \\
                  & EST+PNM & The elder Subhuti, and all the bhikkhus, bhikkhunis, upasakas, upasikas.
\end{tabular}
\end{table*}




Table~\ref{tab:case} displays translation results of three Buddhist scriptures applying four aforementioned models. 

Regarding the first case, baseline and PNM simply translated \begin{CJK*}{UTF8}{gbsn}“色”\end{CJK*} into ``color'' because they have corresponding relationship in general domain. Similarly, baseline and PNM mistranslated the part of speech of \begin{CJK*}{UTF8}{gbsn}“空”\end{CJK*} as the translation only followed the grammatical laws of vernacular Chinese. However, our model translated correctly for proper noun \begin{CJK*}{UTF8}{gbsn}“色”\end{CJK*} and the part of speech of \begin{CJK*}{UTF8}{gbsn}“空”\end{CJK*}.

Concerning the second case, both baseline and PNM incorrectly split \begin{CJK*}{UTF8}{gbsn}“所得法”\end{CJK*} into \begin{CJK*}{UTF8}{gbsn}“所得”\end{CJK*} (income) and \begin{CJK*}{UTF8}{gbsn}“法”\end{CJK*} (method), leading to translation errors as well as negatively affecting subsequent translation. On the opposite, models containing EST translated this scripture correctly because they obeyed linguistic characteristics of vernacular Chinese.

As for the third case, baseline and PNM misregarded \begin{CJK*}{UTF8}{gbsn}“须菩提”\end{CJK*} (a proper noun) as combination of \begin{CJK*}{UTF8}{gbsn}“须”\end{CJK*} (must) and \begin{CJK*}{UTF8}{gbsn}“菩提”\end{CJK*} (another proper noun). Additionally, baseline and EST harnessed transliteration to acquire ``youposai'' and ``youpoy'' while models with PNM obtained ``upasakas'' (proper noun) and ``upasikas'' (proper noun) professionally.

From three cases illustrated above, it should be highlighted that EST can lead to significant improvement in translation coherence and rationality because it draws lessons from vernacular Chinese corpus. What should be equally noted is that PNM can translate proper nouns accurately based on it recall mechanism.

\section{Conclusion}
This paper proposes a novel approach to improve the translation of classical Chinese Buddhist scriptures to English. The problem is a mixture of low-resource translation and domain-specific modeling. Thus, our method combines techniques for the encoder and decoder side to leverage the knowledge from vernacular Chinese auxiliary corpus and raw classical Chinese Buddhist scriptures. The proposed approaches significantly improve the performance on objective and subjective evaluation criteria. As more and more researchers are getting interested in multilingual translation problems, we believe our method can be further generalized to translation scenarios where multiple challenges are encountered.

\section{Acknowledgement}
This paper is supported by the Key Research and Development Program of Guangdong Province under grant No.2021B0101400003. Corresponding author is Jianzong Wang from Ping An Technology (Shenzhen) Co., Ltd (jzwang@188.com).

\bibliographystyle{IEEEtran}
\bibliography{ref}

\end{document}